\documentclass[lettersize,journal]{IEEEtran}
\usepackage{array}
\usepackage[caption=false,font=normalsize,labelfont=sf,textfont=sf]{subfig}
\usepackage{stfloats}
\usepackage{url}
\usepackage{verbatim}
\usepackage{cite}
\usepackage{tabularray}
\usepackage{array}
\usepackage{amsmath,amssymb,amsfonts}
\usepackage{algorithm}
\usepackage{algpseudocode}
\usepackage{booktabs,threeparttable} % added
\usepackage{multirow}
\usepackage{subcaption}
\usepackage{graphicx}
\usepackage{textcomp}
\usepackage{xcolor}
\usepackage{svg}
\usepackage[caption=false,font=normalsize,labelfont=sf,textfont=sf]{subfig}
\usepackage{colortbl}
\usepackage[scientific-notation=true]{siunitx}
\newcommand{\EatOneArg}[1]{}

\begin{document}

\title{PowerGraph-LLM: Novel Power Grid Graph Embedding and Optimization with Large Language Models}

% \author{Fabien~BERNIER, Jun~CAO*, Maxime~CORDY, Salah~GHAMIZI

\author{
    \author{
    \IEEEauthorblockN{Fabien~BERNIER$^{1}$, Jun~CAO*$^{2}$, Maxime CORDY$^{1}$, Salah GHAMIZI$^{2,1}$} \\
    \IEEEauthorblockA{$^{1}$\textit{SnT, University of Luxembourg, Luxembourg}\\
    $^{2}$\textit{Luxembourg Institute of Health (LIH), Luxembourg}\\
    Email: fabien.bernier@uni.lu, jun.cao@list.lu, maxime.cordy@uni.lu, salah.ghamizi@lih.lu}
    }
}

\markboth{}%
{Shell \MakeLowercase{\textit{et al.}}: A Sample Article Using IEEEtran.cls for IEEE Journals}

\maketitle

\begin{abstract}
Efficiently solving Optimal Power Flow (OPF) problems in power systems is crucial for operational planning and grid management. There is a growing need for scalable algorithms capable of handling the increasing variability, constraints, and uncertainties in modern power networks while providing accurate and fast solutions. To address this, machine learning techniques, particularly Graph Neural Networks (GNNs) have emerged as promising approaches. 
This letter introduces PowerGraph-LLM, the first framework explicitly designed for solving OPF problems using Large Language Models (LLMs). The proposed approach combines graph and tabular representations of power grids to effectively query LLMs, capturing the complex relationships and constraints in power systems. A new implementation of in-context learning and fine-tuning protocols for LLMs is introduced, tailored specifically for the OPF problem.
PowerGraph-LLM demonstrates reliable performances using off-the-shelf LLM. Our study reveals the impact of LLM architecture, size, and fine-tuning and demonstrates our framework's ability to handle realistic grid components and constraints.
\end{abstract}


\section{Introduction}
Resolving Alternating Current Optimal Power Flow (AC OPF) problems is a routine task in the operational planning of Power Systems (PS). Nevertheless, with the increasing variability, constraints, and uncertainties in today's power networks, solving problems accurately presents a significant challenge for power system engineers. Numerous strategies for addressing AC OPF challenges incorporate Machine Learning (ML) elements to mitigate the computational challenges associated with traditional optimization-based OPF approaches.

Graph Neural Networks (GNN) have recently demonstrated solid performance for various tasks in the power system. 
In particular, Liu et al.\cite{liu_topology-aware_2023} proposed a new topology-informed GNN approach by combining grid topology and physical constraints. Ghamizi et al. \cite{ghamizi2024hetero} demonstrated that a heterogeneous graph representation combined with physical constraint losses leads to the best performances for PF and OPF tasks. Recent work such as SafePowerGraph\cite{ghamizi2024safepowergraph} provided a standardized representation and benchmark of GNN for PF and OPF problems and identified the best architectures and design choice to solve these problems with GNN.

While these models achieve remarkable performance, they require expensive data curation --- by collecting large training datasets with OPF solvers --- and costly training for specific power grid sizes. Recent progress in foundation models, including LLMs (e.g. ChatGPT, LLaMa), has significantly reshaped the field of machine learning. Such models can indeed generalize to new tasks without expensive training, given the right indications. LLMs have also recently been explored to solve PS-related tasks. \cite{yan2023real} proposed an LLM agent in the Deep Reinforcement Learning (DRL) training loop, directly allowing to model linguistic objectives and constraints in the OPF problem. The optimization is, however, run using traditional methods (solvers and DRL). 
A foundation model, developed in \cite{huang2024large}, can iteratively solve the OPF optimization problem by minimizing the cost function. The optimization only supports the toy example of a simple economic dispatch problem of units and does not fully optimize and predict the OPF variables (generation and bus variables) nor consider real world grid components (multiple loads, line operating limits). To the best of our knowledge, PowerGraph-LLM is the first embedding and optimization framework for OPF that supports realistic grid components and constraints.    

This letter presents three main contributions. We first propose novel power grid embedding for OPF to query LLMs with graph and tabular representations, followed by the development of tailored in-context learning and fine-tuning protocols for these models. We finally include an empirical study on how the architecture, size, and fine-tuning of LLMs affect their performance in solving OPF problems.

\section{PowerGraph-LLM Framework }
\begin{figure*}
    \centering
    \includegraphics[width=0.9\linewidth]{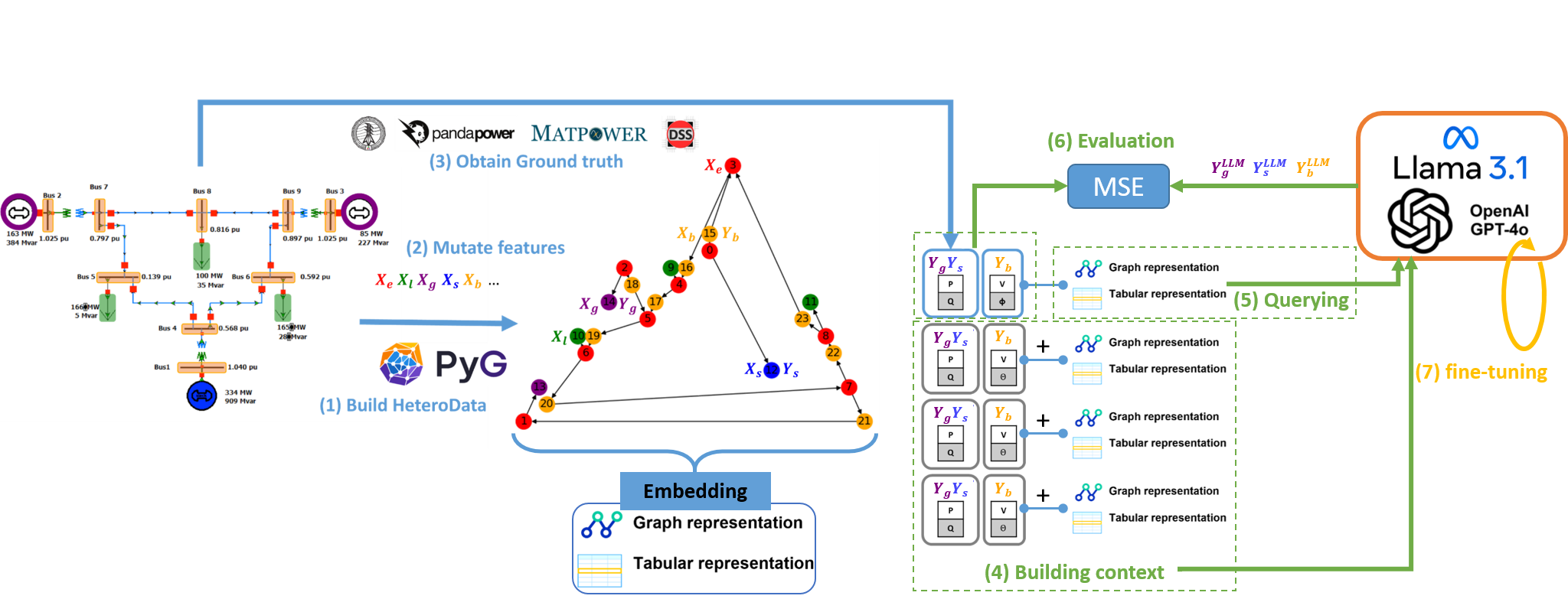}
    \caption{The PowerGraph-LLM framework: We generate the power grid embedding in steps 1, 2, 3 where we obtain a description of the topology of the grid, the features of its components ($X_b$, $X_l$, $X_g$, $X_s$, $X_e$ respectively for bus, loads, generators, slack, and lines), and the OPF solution by a solver ($Y_g$, $Y_s$, $Y_b$). We generate thousands of power grid embeddings and we split them into a context and queries. The context (step 4) is the pairs (power grid description + OPF solutions) that will serve for the LLM model as examples, and the query (step 5) is the grid to which we expect the model to perdict an OPF solution. }
    \label{fig:pipeline}
\end{figure*}

We present in Fig. \ref{fig:pipeline} our PowerGraph-LLM framework. The first three steps consist in building the correct message format to query an LLM for OPF using appropriate table or graph embeddings. The following steps are either implemented using open-source LLMs (Llama), or remote API (OpenAI).

\paragraph{Power Grid Embedding (blue steps 1, 2, and 3 in Fig. \ref{fig:pipeline})}

We extend SafePowerGraph \cite{ghamizi2024safepowergraph} to build the appropriate embedding for LLM.
Starting from an initial grid descriptor (in MatPower, PandaPower or OpenDSS formats), we generate a Pytorch HeteroData with each component as a distinct subgraph (Step (1)). Each node can be of type: bus, load, generator, slack, or line and is associated with its distinct set of features $X_b$, $X_l$, $X_g$, $X_s$, $X_e$ respectively. In order to create new grids, these features are mutated in Step (2) depending on their types. For example, the mutation for loads of the active and reactive power is based on the real profile.

For each mutated grid, Step (3) runs a simulation and solver to derive the PF or OPF optimization's ground truth. This result includes the active and reactive power for every generator and the slack nodes ($Y_g$, $Y_s$), as well as the voltage magnitude and angle for each bus node ($Y_b$).

These features are then prepared for in-context inference in two formats. Our approach can generate LLM messages formatted as \emph{graph representations}, including nodes features and edges connections. We can also generate LLM messages that only contain the features formatted in a tabular format, and we refer to this as the \emph{tabular representation}. Given the context size of an LLM (i.e., how many tokens can be used as input), we can encode more examples in a table representation than in a graph representation.

\paragraph{LLM Inference (green steps 4, 5, and 6 in Fig. \ref{fig:pipeline})}
LLMs predict the next token in a sequence by using contextual information from previously seen words to generate coherent text, facilitated by a chat interface allowing dialogue between a \textit{user} and an \textit{assistant}. The dialogue can be conditioned by a \textit{system prompt}.

A prompt consists in two parts, the \emph{context}, which consists in pairs of examples and expected answers and the \emph{query}, which is the actual input we expect the LLM to predict. In Step (4), the context is pairs of inputs/outputs. The grids (in tabular or graph embedding from the previous step) are the input examples, and the OPF solutions (($Y_g$, $Y_s$, $Y_b$)) are the output examples.
After providing a few examples, we query the LLM in step (5) with the actual grid to predict its OPF solution. The query only consists in the grid embedding either in tabular or graph representation without the OPF solutions.

The whole discussion consisting in the system prompt, the context and the query is processed in text format by the LLM.
GPT and Llama models~\cite{grattafiori2024llama3herdmodels}, two state-of-the-art LLMs families we consider in this paper, share a common high-level architecture --- consisting in breaking text into tokens, converting them to embeddings, and processing them through transformer blocks to capture data patterns~\cite{brown2020languagemodelsfewshotlearners}.

\paragraph{LLM fine-tuning with LoRA}
The Low-Rank Adaptation (LoRA)~\cite{hu2021loralowrankadaptationlarge} method allows fine-tuning LLMs efficiently by introducing low-rank matrices that capture task-specific adaptations while keeping the main model weights unchanged.
Formally, instead of updating all the weights, LoRA modifies a small set by weight update $\Delta W$ as:
\begin{equation}
    \Delta W = \frac{\alpha}{r} A \cdot B
\end{equation}

where $A \in \mathbb{R}^{d \times r}$ and $B \in \mathbb{R}^{r \times k}$, with $r \ll d, k$ and $\alpha$ being a chosen scaling factor. Only $A$ and $B$ are trained, reducing computation and memory needs, allowing for efficient adaptation to new tasks with few data and lower expenses while maintaining the LLMs' pre-existing capabilities.

We run the fine-tuning process in step (7) in yellow in Fig. \ref{fig:pipeline}. We consider each input example as a separate training sample and minimize the learning loss to its associated OPF solution.

\section{Empirical Study}
\begin{table*}[!ht]
    \centering
    \caption{Errors in OPF estimation after fine-tuning GPT4o-mini and Llama-8b.}
    \label{tab:finetuning}
    \begin{tabular}{llrrrr|rrrr}
    \toprule
                      Case & LLM & \multicolumn{4}{c|}{Before fine-tuning}                                                                                       & \multicolumn{4}{c}{After fine-tuning}                                                                                        \\ 
                       
                       & & $MSE_{GEN}$ & $MSE_{SLACK}$ & $MSE_{BUS}$ & VALID & $MSE_{GEN}$ & $MSE_{SLACK}$ & $MSE_{BUS}$ & VALID \\ \cline{1-10}
    9-bus & GPT4o-mini (graph) & \num{0.205558}                     & \num{0.177675}                       & \num{0.000806}                     & 97.7\%                            & \num{0.018740}                     & \num{0.010669}                       & \num{0.000269}                     & 93.1\%                            \\
    & Llama-8b (graph)   & \num{8461773.406823}               & \num{2213.237469}                    & \num{0.074695}                     & 11.5\%                           & \num{0.006417}                     & \num{0.061813}                       & \num{0.000629}                     & 99.7\%                             \\
    & Llama-8b (table)   & \num{3.172312}                     & \num{1.957585}                       & \num{0.001922}                     & 32.1\%                           & \num{0.054423}                     & \num{0.052320}                       & \num{0.001846}                     & 98.4\%            \\
    30-bus & Llama-8b (graph)   & -               & -                    & -                     & 0\%                           & \num{0.01807}                     & \num{0.016839} & \num{0.000513} & 89.5\%                             \\
    \bottomrule
\end{tabular}

\end{table*}

\subsection{Experimental Protocol}
\paragraph{In-context inference}
To investigate the capability of LLMs in generalizing from examples presented within the context window, we conduct an assessment using in-context inference. Four models were tested: OpenAI's \textit{gpt-4o-mini}, OpenAI's \textit{gpt-4o}, \textit{Llama-3.1-8B-Instruct}, and \textit{Llama-3.1-70B-Instruct}~\cite{grattafiori2024llama3herdmodels}, the latter being respectively renamed \textit{llama-8b} and \textit{llama-70b} for the sake of brevity.
The inference for \textit{llama} models is performed using \textit{ollama} v0.5.0, with its default setup, including a context window of 4096 tokens.

Sequences provided for in-context inference are made as follows: after an initial system prompt, a total of 65 pairs of example requests and solutions are provided using the JSON format. A context of 65 examples maximizes the utilization of context windows across all models.
An additional, 66$^\text{th}$ example is used to evaluate the model's generalization abilities, with its response benchmarked against an expected solution. The overall sequence constructed is illustrated in Table \ref{tab:llm-sequence}.

\begin{table}[!ht]
    \centering
        \caption{Sequence schema provided for LLM inference}
    \begin{tabular}{ p{8cm} }
        \toprule
        \textbf{system:} \\
        You are a power grid operator running an Optimal Power Flow simulation, and you need to return a JSON-formatted response based on the provided input JSON. The input is the description of the components of the grid, including the buses, generators, loads, lines, and external grid. The output is the solution to the optimal power flow problem. You will get a few examples of Input and Output JSON. You need to return the correct Output for the last given Input. \\
        \hline
        \textbf{user:} \\
        {Example Input JSON}: \textless{embedding input \#1}\textgreater \\
        \hline
        \textbf{assistant:} \\
        {Example Output JSON}: \textless{OPF solution \#1}\textgreater \\
        \hline
        \hspace{3.8cm} . . . \\
        \hline
        \textbf{user:} \\
        {Query Input JSON}: \textless{embedding input \#66}\textgreater \\
        \bottomrule
    \end{tabular}
    \label{tab:llm-sequence}
\end{table}

If no JSON object can be read from the LLM's response, or if the values returned by the LLM are invalid (missing or invalid values), the output is deemed \textit{INVALID}.
This evaluation process was repeated 1,000 times, resulting in the generation of a total of 65,000 pairs for context and 1,000 pairs for evaluation across the assessments.

We run these inference tasks on one NVIDIA RTX 8000 48GB for Llama 8B (two for Llama 70B), using Q4\_K\_M quantization
 to optimize performance and resources.

\paragraph{Fine-tuning}
Subsequently, we apply fine-tuning to the \textit{Llama-3.1-8B-Instruct} model, as shown on Figure \ref{fig:pipeline}, using the 65,000 pairs of context developed earlier. Each fine-tuning sample includes the system prompt, the specific OPF problem being addressed, and its corresponding solution. We follow a similar protocol to fine-tune OpenAI's \textit{gpt-4o-mini} with API. 

The fine-tuning process of the \textit{Llama} model was conducted on an NVIDIA RTX 8000 48GB, using a LoRA configuration. The LoRA setup was defined with a rank of 8 ($r = 8$) and a scaling factor of 16 ($\alpha = 16$). Both fine-tuning and inference of the fine-tuned model were performed using the Hugging Face library.

\paragraph{Evaluation}

In all experiments, we evaluate the mean squared error (MSE) for the active and reactive powers of the generators and the slack bus, as well as the voltage magnitude and phase angle of the buses.
Each test grid is an IEEE 9-bus grid where the loads are mutated following a uniform distribution (+/- 20\% variations).
We also report the percentage of test grids where the output of the LLM was \textit{invalid}, i.e. no JSON could be parsed from the output. This typically happens when the assistant tries to explain how to solve the problem instead of providing the solution itself.
Reported values include MSE for generators (GEN), slack buses (SLACK), and buses (BUS) values in the grid.

\subsection{Effectiveness of pre-trained LLMs}

\begin{figure}[!ht]
    \centering
    \includegraphics[width=\linewidth]{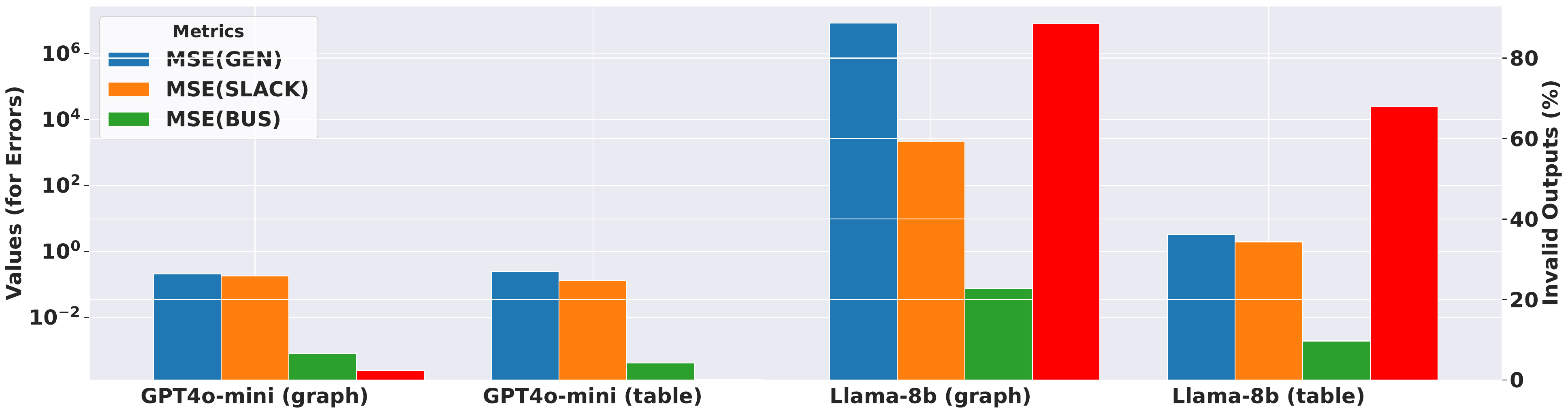}
    \caption{Errors in OPF estimations for \textit{gpt-4o-mini} and \textit{llama-8b}. The red bars represent the proportion of invalid inputs.}
    \label{fig:small-results}
\end{figure}

\paragraph{ChatGPT (proprietary) vs. Llama (open-source)}
The results, presented in Figure~\ref{fig:small-results}, clearly show an important discrepancy between \textit{gpt-4o-mini} and \textit{llama-8b}. While \textit{gpt-4o-mini} consistently showcases GEN and SLACK MSE values below 0.5 and BUS MSE below $10^{-3}$ for both graph and table formats with minimal invalid outputs, \textit{llama-8b} exhibits higher MSEs by multiple magnitudes (min. 100 times more) for all criteria and a large majority of invalid outputs.
\paragraph{Graph vs. tabular representations}
The usage of the tabular representation here results in lower errors than graphs for both models, although it is less pronounced for the GEN and SLACK with \textit{gpt-4o-mini}. Table representations also lead to less invalid outputs by 23\%, supporting their efficiency.

\subsection{Impact of the size of the model}
\begin{figure}[!ht]
    \centering
    \includegraphics[width=\linewidth]{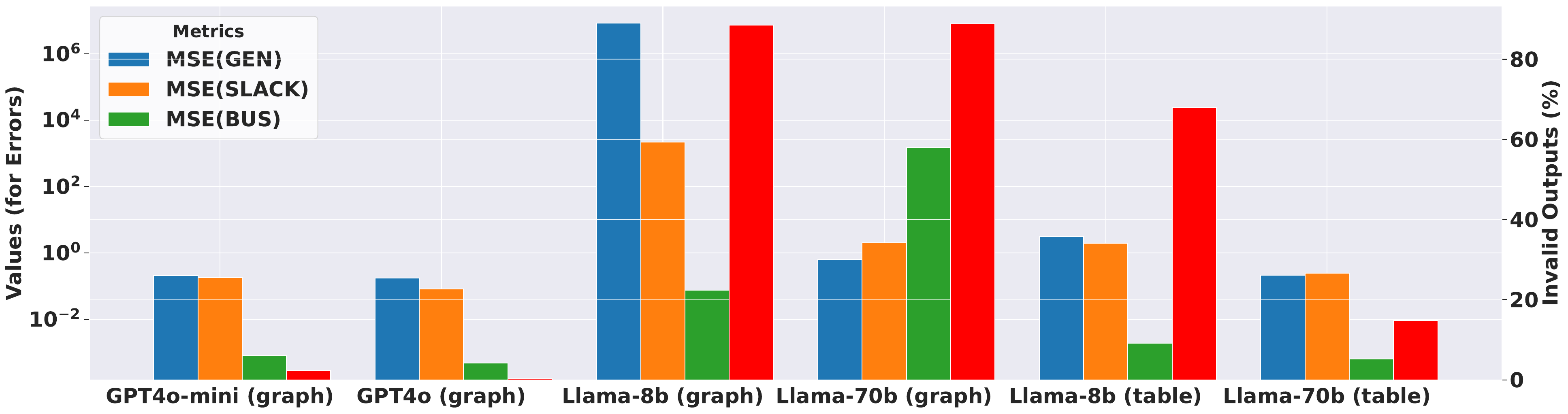}
    \caption{Errors in OPF estimations for graph representations using bigger \textit{gpt-4o} and \textit{llama} models. The red bars represent the proportion of invalid inputs.}
    \label{fig:size-results}
\end{figure}

As it is the case for natural language tasks~\cite{grattafiori2024llama3herdmodels}, bigger models result in better performance (Figure \ref{fig:size-results}). This improvement is especially prominent for \textit{llama-70b}, getting closer to \textit{gpt-4o-mini}'s performance in terms of GEN and SLACK MSE.

The MSE of BUS decreases when increasing the size of the LLM using table representation but increases significantly with graph representation. 
Our results confirm that larger LLMs lead to better performance for given representations, and both the size and the representation parameters should be considered together in future assessments.

\subsection{Impact of fine-tuning}
We report in Table \ref{tab:finetuning} the impact of fine-tuning LLMs to solve the OPF problem. The percentage of invalid outputs of the LLM decreases to less than 2\% for the Llama models and marginally increases for the GPT4o model. The error across all the components decreases, in particular for the Llama models. They become as effective as the proprietary model. 

Contrary to the earlier vanilla models, graph representation is more effective than tabular to query LLMs after fine-tuning. 

\begin{figure}[!ht]
    \centering
    \includegraphics[width=0.9\linewidth]{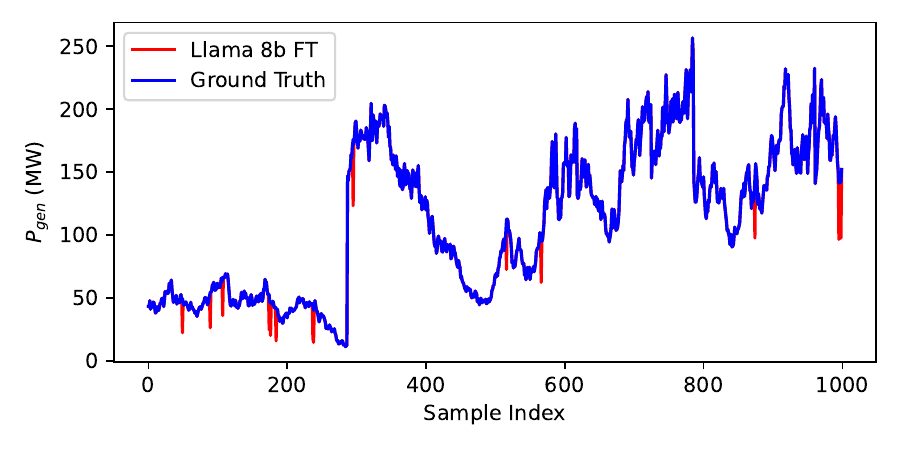}
    \caption{Predicted and real value (Gen 1) with Llama-8b.}
    \label{fig:qualitative-results-gen}
    \includegraphics[width=0.9\linewidth]{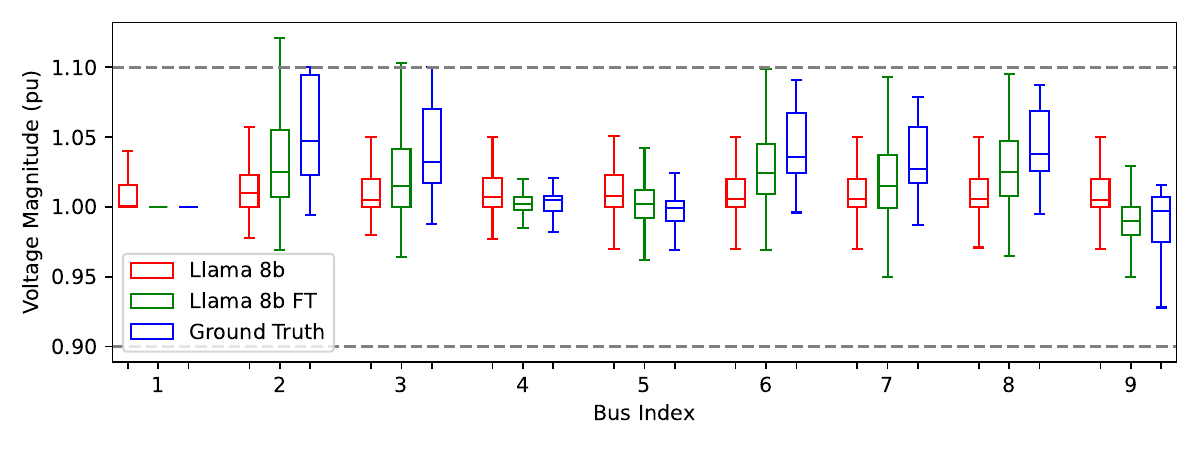}
    \caption{Predicted and real voltage of buses with Llama-8b.}
    \label{fig:qualitative-results-bus}
\end{figure}

In Fig. \ref{fig:qualitative-results-gen}, we compare the predicted active power of the generator and the ground truth. After fine-tuning, the model achieves faithful results within large ranges of perturbations. Similarly, Fig. \ref{fig:qualitative-results-bus} shows that the fine-tuned LLM leads to closer results while achieving only rare violations (Bus 2).

We focus on the graph representation to evaluate the generalization of larger grids. Our results on the 30-bus grid in Table \ref{tab:finetuning} demonstrate that off-shelf LLMs completely fail to generate valid OPF solutions, but that LLMs can be fine-tuned to achieve competitive results with graph embedding.

\section*{Conclusion}
Our letter introduces PowerGraph-LLM, a novel framework for solving OPF problems using LLMs. We propose a new power grid embedding combining graph and tabular representations, LLM fine-tuning protocols for OPF, and an empirical analysis of LLM performance. Our results show that larger models perform better, with fine-tuning significantly improving accuracy and reducing invalid outputs. Graph representations become more effective than tabular ones after fine-tuning. 
These findings highlight the potential of LLMs in power system optimization and open new avenues for more efficient and accurate OPF solutions for complex power grids.

\section*{Acknowledgements}
This work was partially funded by FNR CORE project LEAP (17042283) and by Creos Luxembourg S.A. \\
Dr. Ghamizi is supported by the Luxembourg National Research Fund (FNR) CORE C24/IS/18942843.

% {
% \bibliography{bib/leap,bib/ml,bib/other}
% }
% \bibliographystyle{IEEEtran}

\newpage

\vfill

\end{document}